%% file: main.tex
\documentclass[runningheads]{llncs}

\usepackage{eccv}

\usepackage{eccvabbrv}

\usepackage{graphicx}
\usepackage{booktabs}

\usepackage[accsupp]{axessibility}  

\usepackage[pagebackref,breaklinks,colorlinks,citecolor=eccvblue]{hyperref}

\usepackage{orcidlink}

\usepackage{colortbl}
\usepackage{tikz}
\usepackage{multirow}
\usepackage{makecell}
\usepackage{bbding} 
\usepackage{pifont}
\usepackage[mathscr]{eucal}
\usepackage{bm}
\usepackage{float}
\newcommand{\xmark}{\ding{55}}
\definecolor{darkF7E0D5}{RGB}{209,154,128}
\newcommand{\rownumber}[1]{\textcolor{darkF7E0D5}{#1}}
\definecolor{aliceblue}{rgb}{0.94, 0.97, 1.0}

\newcommand{\tablestyle}[2]{\setlength{\tabcolsep}{#1}\renewcommand{\arraystretch}{#2}\centering\footnotesize}

\begin{document}

\title{Merlin: Empowering Multimodal LLMs with Foresight Minds} 

\titlerunning{Merlin}

\author{En Yu\inst{1}\thanks{Equal Contribution} \and
Liang Zhao\inst{2\star} \and
Yana Wei\inst{3} \and Jinrong Yang\inst{3} \and Dongming Wu\inst{4} \and Lingyu Kong\inst{5} \and Haoran Wei\inst{2} \and Tiancai Wang\inst{2} \and Zheng Ge\inst{2} \and Xiangyu Zhang\inst{2} \and Wenbing Tao\inst{1}\thanks{Corresponding Author}}

\authorrunning{Yu, E. et al.}

\institute{Huazhong University of Science and Technology \and
MEGVII Technology \and ShanghaiTech University \and
Beijing Institute of Technology \and University of Chinese Academy of Sciences
\\
\email{\{yuen,wenbingtao\}@hust.edu.cn}}

\maketitle

\begin{figure}[H]
\centering
\hsize=\textwidth
    \includegraphics[scale=0.61]{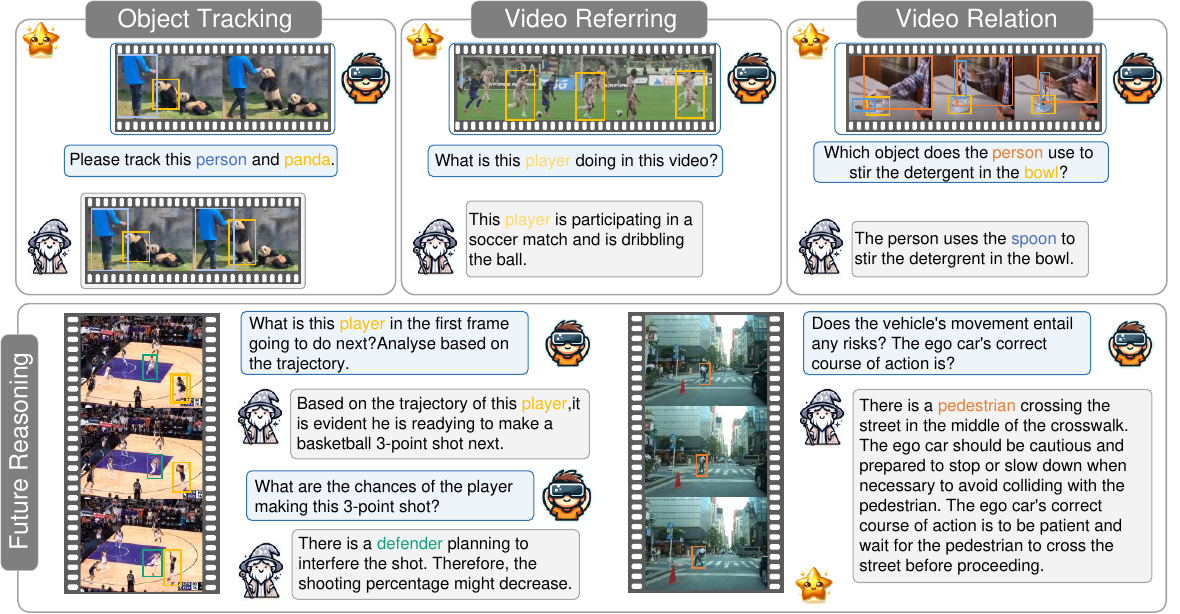}
    \vspace{-2mm}
    \vspace{-2mm}
    \caption{\textbf{Demo cases presentation of Merlin}. Here we showcase several main capabilities of our built Multimodal Large Language Model (MLLM). Notably, in the dialogue, the words marked with colors correspond to the trajectory outputs of the targets in the image. To save space, we highlight them using the same colors.}
    \label{fig:examples}
\end{figure}

\begin{abstract}

Humans can foresee the future based on present observations, a skill we term as \textit{foresight minds}. However, this capability remains under-explored within existing MLLMs, hindering their capacity to understand intentions behind subjects. To address this, we integrate the future modeling into MLLMs. By utilizing the \textbf{trajectory}, a highly structured representation, as a learning objective, we aim to equip the model to understand spatiotemporal dynamics. Inspired by the learning paradigm of LLMs, we first propose \textit{Foresight Pre-Training (\textbf{FPT})} that jointly learns various tasks centered on trajectories, enabling MLLMs to predict entire trajectories from a given initial observation. Then, we propose \textit{Foresight Instruction-Tuning (\textbf{FIT})} that requires MLLMs to reason about potential future events based on predicted trajectories. Aided by FPT and FIT, we build an unified MLLM named \textbf{\textit{Merlin}} that supports complex future reasoning. Experiments show Merlin’s foresight minds with impressive performance on both future reasoning and visual comprehension tasks. Project page: \url{https://ahnsun.github.io/merlin}.

\end{abstract}

\section{Introduction}
\label{intro}

Human beings can predict future events or outcomes based on current observations, known in neuroscience theory as \textit{predictive processing}~\cite{friston2010free}. In this paper, we refer to this ability as \textit{foresight minds}, which involves the use of past experiences, knowledge, sensory information, and probabilistic reasoning to generate expectations about future events. In the artificial intelligence (AI) domain, the capability to predict future events is an important topic towards the realization of artificial general intelligence (AGI).

\begin{figure}[t]
\centering
\includegraphics[width=0.95\linewidth]{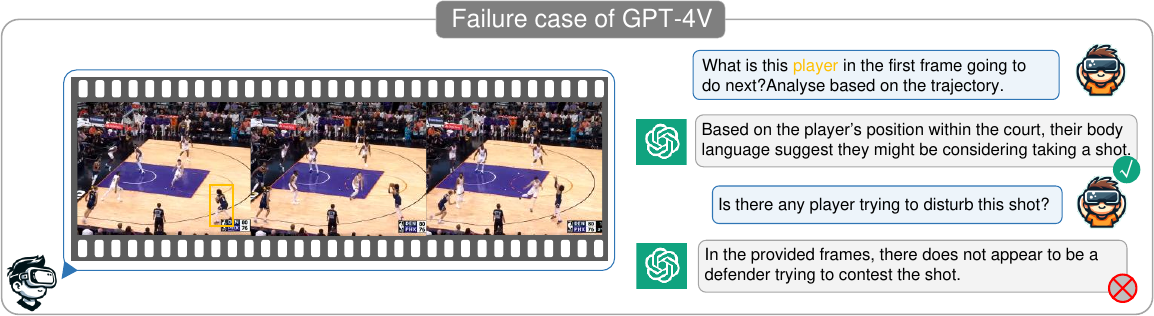}
\caption{\textbf{Failure case of GPT-4V} about future reasoning.}
\label{fig:fail}
\end{figure}

Recent advancements in Multimodal Large Language Models (MLLMs), such as GPT-4V~\cite{GPT4} and Bard~\cite{anil2023palm}, have shown significant potential in image understanding and logical reasoning. Despite these achievements, these models struggle to foresee future events based on current image observations. Even provided with additional observations, like sequences of multiple frames, the current MLLM models still struggle to adequately analyze and infer specific target behaviors, such as predicting object movements or interactions (shown in Figure~\ref{fig:fail}). On the contrary, human can reason the future to some extent based on the observed current state~\cite{human,huamn1}, which shows powerful foresight minds. 

To mitigate this existing deficiency in MLLMs, we start from dividing human's process of 
foreseeing the future into a two-stage system~\cite{human,human2}: (1) observing the dynamic clues of the subject and then (2) analyzing the behavior pattern and reasoning what might happen according to the observation. For instance, while watching a basketball game, people will first observe the moving players on the court, and then forecast the specific player's forthcoming actions, e.g., shooting, slam-dunking, or passing, by analyzing the current states and movement patterns of the players. Compare this system to current MLLMs, we find that MLLMs can complete the second stage well, thanks to the powerful logical reasoning ability of LLM~\cite{wei2022chain, pi2023detgpt}. Therefore the key challenge is the first stage. That is, \textit{how to make MLLM acquire correctly \textbf{spatiotemporal dynamics} from the multi-image observation?}   

Explicitly modeling next frames (\emph{e.g.}, reconstructing next frames~\cite{yan2021videogpt,cholakov2021transformers}) can be a straightforward way. However, it can be hard to directly extract dynamic clues from the redundant visual information~\cite{he2022masked}, especially from video sequences. It is necessary to construct a suitable learning objective to assist MLLM in obtaining dynamic clues about the specific subjects. To this end, we point out that \textit{\textbf{trajectory}, as a highly structured representation, is a good learning objective which can link the temporal contexts between the past and the future.}

Based on this insight, we propose to model the future to empower existing MLLMs with ``\textit{foresight minds}''. Following the modern learning paradigm of LLMs, our future reasoning learning process includes two stages: (1) \textit{Foresight Pre-Training (\textbf{FPT})}, a paradigm that causally models the temporal trajectories, which interleave with multi-frame images. The model starts with the initial observation of one or multiple subjects in the first frame as the query and then is required to predict the whole trajectory. Notably, we introduce various tasks containing richly labeled data~\cite{objects365,fan2019lasot,huang2019got,sun2022dancetrack,refcoco,vcr}, including object detection, object tracking, etc., to perform multitask learning. And samples from these tasks are properly formatted to ensure coordinated pre-training.
(2) \textit{Future Instruction-Tuning (\textbf{FIT})}, then, considers the trajectory modeling bestowed by FPT as a bridge in the logical chain of future reasoning. Simply put, when querying an MLLM, it must articulate its reasoning in conjunction with the trajectory for each object referenced. This method, as a form of \textit{Trajectory Chain-of-Thought}, effectively narrows the gap between trajectory perception and predictive future reasoning, thereby fully unleashing model's foresight minds.

Aided by the above future modeling technologies, we provide \textit{\textbf{Merlin}}\footnote{\textbf{Merlin} is a legendary character in the tales of King Arthur, renowned as a powerful wizard and a wise counselor in the Arthurian legends. He is depicted as having the power to foresee future events and has a deep understanding of fate and destiny.}, a novel and unified MLLM capable of handling inputs and outputs of spatial coordinates or tracklets from single image or multiple frames. 
Moreover, Merlin is adept at performing inductive reasoning about future events based on current observational results. To demonstrate this, we provide several real dialogues between users and Merlin, as displayed in the Figure~\ref{fig:examples}. Unlike the previous MLLMs~\cite{llava, minigpt4, chatspot} which only supported interaction with a single image, Merlin not only provides a richer multi-image interaction, but also on this basis, is capable of executing unique and powerful future reasoning.


We construct a new future reason benchmark to evaluate Merlin's logical reasoning and future prediction abilities. The results, which significantly surpass previous baselines~\cite{shikra,llava,llava1.5,ye2023mplug}, demonstrate Merlin's stunning performance in future reasoning.
We further reveal Merlin's exceptional performance in general visual understanding. Through analysis in scenarios such as VQA (Visual Question Answering)~\cite{VizWiz,hudson2019gqa}, comprehensive understanding~\cite{liu2023mmbench, yu2023mmvet}, and hallucination~\cite{li2023pope}, we unexpectedly discovered that our proposed novel paradigm of future learning aids MLLMs in gaining a deeper understanding of images. We believe this brings new insights for the training of future MLLMs.

\section{Related Work}
\label{releated work}
\subsection{Large Language Models}
Large Language Models (LLMs) have gained significant attention due to their capabilities in language generation and logical reasoning. Pioneering models like BERT~\cite{Bert}, GPT-2~\cite{GPT-2}, and T5~\cite{T5} laid the groundwork, but GPT-3~\cite{GPT3}, the first model with a 175 billion parameter size, made notable strides, demonstrating strong zero-shot performance. An emergent ability, wherein model size scaling results in significant language capability improvements, was also observed in LLMs. This was further facilitated by InstructGPT~\cite{InstructGPT} and ChatGPT~\cite{ChatGPT} using Reinforcement Learning with Human Feedback (RLHF) on GPT-3. These advancements led to what's called LLMs' ``iPhone moment''. Following GPT's success, several open-source LLMs, including OPT~\cite{OPT}, LLaMA~\cite{llama}, and GLM~\cite{GLM}, have been proposed, showing similar performance to GPT-3. Models like Alpaca~\cite{alpaca} and Vicuna~\cite{vicuna} illustrate the application of these LLMs, using a self-instruct framework to construct excellent dialogue models.

\subsection{Multimodal Large Language Models}

The advancements in LLMs~\cite{ChatGPT, llama, llama2} have projected a promising path towards artificial general intelligence (AGI). This has incited interest in developing multi-modal versions of these models.
Current Multi-modal Large Language Models (MLLMs) harness the potential of cross-modal transfer technologies. These models consolidate multiple modalities into a unified language semantic space, and then employ autoregressive language models as decoders for language outputs. Models like Flamingo~\cite{Flamingo} have adopted a gated cross-attention mechanism, trained on billions of image-text pairs, to align visual and linguistic modalities, showing impressive performance on few-shot learning tasks. Similarly, BLIP-2~\cite{BLIP2} introduced the Q-Former to align visual features more effectively with language space. The LLaVA series~\cite{llava, llava1.5} further enhanced this process by using simply a MLP in place of the Q-Former and designing a two-stage instruction-tuning procedure. Apart from creating general MLLMs, techniques have also been developed for visual-interactive multimodal comprehension, involving the precise tuning of referring instructions~\cite{zhao2023chatspot, shikra, zhang2023gpt4roi}. Furthermore, another interesting direction in MLLM research involves integrating MLLMs for cross-modal generation~\cite{dong2023dreamllm, koh2023generating, ge2023planting} by using text-to-image models such as Stable Diffusion.

\begin{figure*}[t]
\centering
\includegraphics[width=1.0\linewidth]{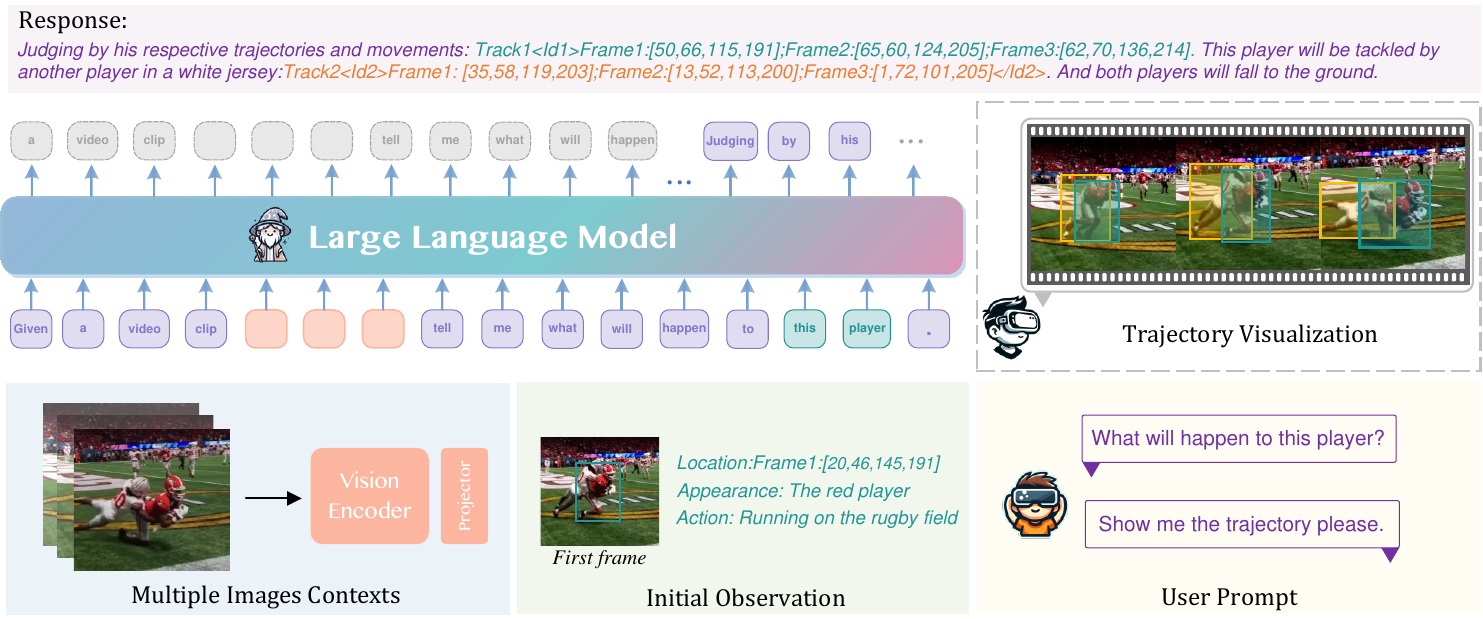}
\vspace{-4mm}
\caption{\textbf{Overall pipeline of Merlin}. The architecture of Merlin consists of three main components: (1) an image encoder, (2) a large language model, and (3) a modality-align projector. \textbf{Bottom:} The diverse input format that supports multiple-image contexts, initial observation and the specific user prompt. \textbf{Top:} The model response including the predicted trajectory and the future reasoning.}
\label{fig.3}
\end{figure*}

\section{Metholodgy}
\label{method}

In this section, we introduce the design details of Merlin, encompassing its overall architecture (Section~\ref{overall}), and the two-stage training pipeline of foresight learning, including Foresight Pre-Training (Section~\ref{FPT}) and Foresight Instruction Tuning (Section~\ref{FIT}).

\subsection{Overall Architecture}
\label{overall}

Merlin is designed to unlock the foresight minds based on observations from single images and multi-frame video clips. In order to accomplish this, images and videos are comprehensively represented through a series of visual tokens, which are then integrated into the language sequence that can be comprehended by Large Language Models (LLMs) in a unified framework. Specifically, Merlin consists of an image encoder, a decoder-only LLM, and a modality alignment block as illustrated in Figure \ref{fig.3}. Following prevalent practice~\cite{llava,llava1.5,minigpt4,shikra}, we opt for the pre-trained CLIP~\cite{clip} ViT-L/14~\cite{vit} as the visual encoder and Vicuna-7B v1.5~\cite{vicuna} as the large language decoder. For more details, please refer to our supplementary materials.

To provide enough visual information and details, the input images are resized to a resolution of $448 \times 448$. At this juncture, the visual encoder iteratively attends to $(448 / 14)^2$ uniformly divided image patches, yielding $1024$ encoded tokens. Considering the limited context length of LLMs and addressing the substantial computational challenges posed by high resolution and multi-frame context modeling, we simply utilize a 2D convolution to achieve both dimension projection and token aggregation~\cite{tokenmerge,tokenlearner}. 

We choose 2D convolution over 1D linear layers~\cite{llava,llava1.5,shikra} or cross-attention layers~~\cite{BLIP2,minigpt4,qwen} as connector for the following reasons: (1) 2D convolution clusters local visual tokens on a spatial scale~\cite{conv_1}, effectively achieving a one-step transformation from spatial to channel information; (2) The good convergence properties~\cite{bn,vgg} of 2D convolution compared with cross-attention lay a solid foundation for foresight learning in a two-step training approach.

\subsection{Foresight Pre-Training}
\label{FPT}

Generative Pre-Training (GPT)~\cite{GPT3,ChatGPT,GPT4} serves as the cornerstone of this generation's Language Models (LLMs). Through learning to predict next token, the model efficiently condenses data, thereby yielding emergent forms of intelligence~\cite{emegen}. In this context, a very natural approach to enhance the model's perception of the dynamic clues across multiple frames is to \textit{explicitly model the next frame} (or image). However, due to the high redundancy in multi-frame visual information, the truly next-frame prediction remains a significant challenge to date. A better approach at this juncture is to \textit{implicitly}\textit{ model high semantic information in the label space} (such as categories, bounding boxes) on a frame-by-frame basis. Temporally, this label information forms a \textit{\textbf{trajectory}}, a highly structured representation. Causally modeling the trajectory in conjunction with each frame of image helps to connect the past and present in time, thus enabling the model to perceive the future.

To this end, we propose the Foresight Pre-Training, a way of \textit{causally modeling the trajectories interleaved with multi-frame images}, to empower the MLLM with the capacity of perceive the dynamic clues, and ultimately achieving future reasoning. Specifically, given a video clip including several frames, we first give the model the observation of the \textit{first frame}, then we require the model to predict the \textit{whole trajectory} of the concerned subject in this video conditioned on the initial observation. Notably, the observation of the first frame can be the description or simple position of the concerned object. Formally,
\begin{equation}
\begin{aligned}
P(Y|X) \sim P(Y|\{X_1,X_2,...\}, O_{first}),
\label{eq:pnf}
\end{aligned}
\end{equation}
where $X_{i}$ denotes the $i^{th}$ frame and $O_{first}$ is the first frame observation, $Y$ refers to the trajectory of the subject in $O_{first}$ within the frame sequence. The observation and the raw frames will be regarded as the condition to prompt MLLM to predict the trajectory.

\noindent\textbf{Data Construction.} 
We first aggregate all valuable multimodal information from diverse data resources and then properly organize them for multi-task foresight pre-training.
Specifically, for each sample instance $\mathbf{I}$, we first collect its multimodal information including consecutive multi-frame images $\{X_1,X_2,...\}$, subject observations from the \textit{first} frame $O_{first}$, and subject trajectory $Y$ constructed from \textit{all} frames. Formally,
\begin{equation}
\begin{aligned}
\mathbf{I} = \mathbf{\{}\{X_1,X_2,...\}, O_{first},Y \mathbf{\}}.
\label{eq:data_format}
\end{aligned}
\end{equation}
We categorize observations of one subject of the first frame into three types: location description, appearance description and action description. Then we \textit{randomly} select one of these observations of a particular subject in the first frame as the query object. 

\begin{figure}[t]
\centering
\includegraphics[width=1.0\linewidth]{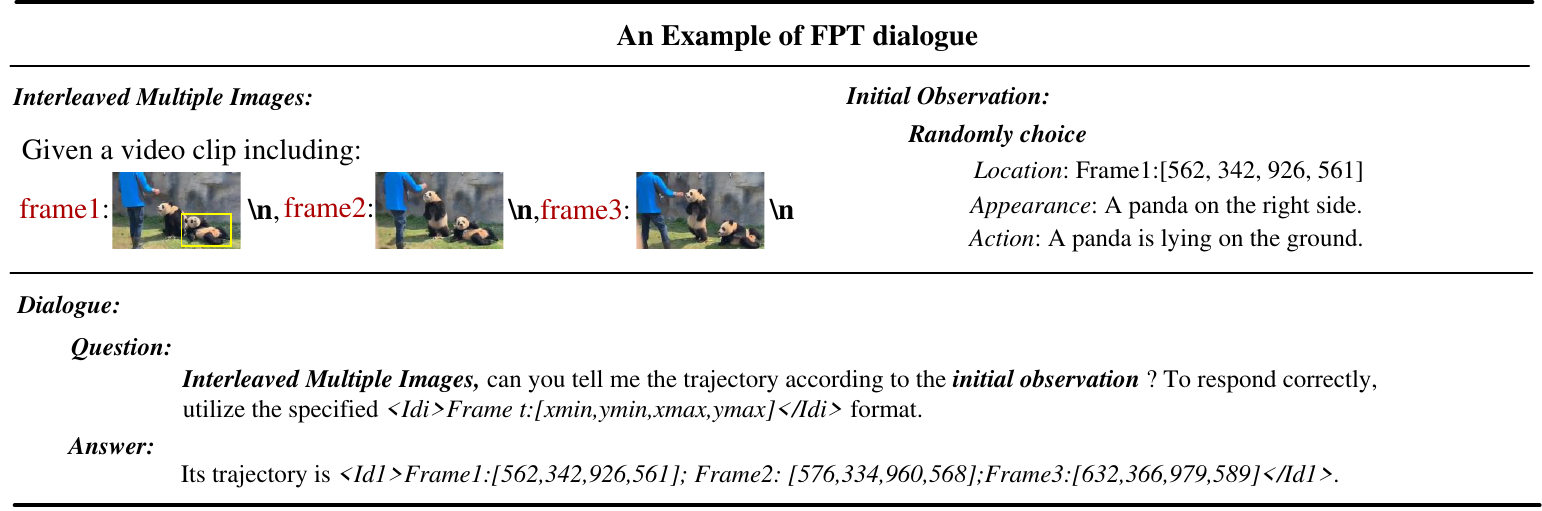}
\caption{\textbf{One example to illustrate the multi-modality pretraining dataset}. The top block shows the provided contexts including the multiple images contexts and initial observation (box, appearance and action) about the subject to prompt the LLM. The bottom block shows the dialogue including question and answer.}
\label{fig.4}
\end{figure}

To better unleash the powerful generative modeling capacity of LLM, we construct this query process as a type of conversation. Here is an example of the constructed data shown in Figure~\ref{fig.4}.  In this case, we want to query the subject --- the panda on the right --- with the randomly select observation, and expect the answer with the movement trajectory of this panda across multiple frames. To model this process, we convert the query to question and trajectory to answer with proper natural language for embellishment.

Overall, the aforementioned process of dialogization roughly follows these three principles: (1) \textbf{Precise definition of task prompts and answer formats}. In particular, we use a task prompt to tell MLLM \textit{what specific task to do} (detect or track), and also \textit{specified the answer format} with accurate descriptions in each question. In this way, different types of tasks can be flexibly organized together without compromising the general language ability. (2) \textbf{Clear indication of multimodal information.} Concretely, for each group of image tokens, we add a \textit{special frame indicator} in front of then, i.e., \textit{frame1:\texttt{<image>} and frame2:\texttt{<image>}}, so as to help MLLM better focus on the corresponding image. 
(3) \textbf{Interleaving of frames and observations.} For the same identity, we interleave the frames in which it appears with its positional observations, and enclose them with two ID tokens (i.e. \texttt{<Idi>} and \texttt{</Idi>}) to construct a trajectory. 
We believe that this interleaved organization helps in \textit{generatively training to model causality within the trajectory}, while the ID tokens ensures that the model can distinguish among different identity objects. 

\noindent\textbf{Training Details.} 
The objective in this stage is to initially endow MLLMs with the capacity of modeling the spatiotemporal dynamics across multi-frame images, while ensuring that its general language capabilities do not diminish.
Previous practices~\cite{llava, qwen, liu2023improved} typically conducting a separate modality alignment training phase following a multi-task pre-training stage, which however, complicates the training process and data construction. In this paper, we directly incorporate both of them into one stage, and \textit{unfreeze all modules during pre-training}. This is because that we believe the MLLMs are sufficiently powerful to concurrently handle the learning of general multimodal capabilities and multi-task specific abilities \textit{under proper guidance}. 
Furthermore, we mix a large amount of image-text pairs and rich-annotated conversation data (formatted according to the above method) from diverse data sources~\cite{fan2019lasot,huang2019got,milan2016mot16,simsek2022sompt22,sun2022dancetrack,refcoco,shao2019objects365, vcr} to conduct multi-task learning. 
In doing so, not only endows the model with foresight minds but also ensures its multimodal alignment.

\subsection{Foresight Instruction Tuning}
\label{FIT}

Althought Foresight Pre-Training equips the model with the ability to observe dynamic clues across multiple frames, it still falls short of true foresight minds. This is because models typically struggle to effectively transform such observations into successful future reasoning~\cite{tom,tom2}. 

Recent work~\cite{pi2023detgpt,tom} has highlighted that Chain-of-Thought (CoT)~\cite{wei2022chain} is crucial in bridge the gap between the observations and actions of MLLMs with theory of mind~\cite{tom1,tom2}. Meanwhile, several prior studies~\cite{shikra,chatspot} have also demonstrated that prompts indicating position (such as bounding boxes or points) --- a principle analogous to CoT --- can concentrate an MLLM's attention on the relevant area, leading to more accurate dialogues and reducing the likelihood of visual hallucination.
Drawing inspiration from these findings, we conduct the Foresight Instruction Training (FIT) building upon the foundation of FPT to further enhance the model's future reasoning capability. In specific, building on the trajectory generating powered by FPT, we further union the trajectories to generatively rationalize the forthcoming events. Mathematically,
\begin{equation}
\begin{aligned}
P(Z|X,Y) \sim P(Z|\{X_1,X_2,...\}, O_{first}, Y),
\label{eq:FPT}
\end{aligned}
\end{equation}
where $Z$ refers to the future observation which is deduced from observations in \textit{each} frame. It can be actions, events, trends, or simply likelihoods. In this context, multi-frame images, in conjunction with the first subject observation, and the trajectory of the same subject across all frames, serve as the union condition to prompt MLLM to causally predict the future. This way, akin to a \textbf{\textit{Trajectory Chain-of-Thought}}, effectively bridges the gap between trajectory perception and predictive future reasoning, thereby fully unleashing model's foresight minds.

\noindent\textbf{Data Construction.} 
The specific data construction method is similar to FPT, but on this basis, we also deduce a future observation $Z$ from the information across multiple frames and append it after the trajectory in the answer. Formally, 
\begin{equation}
\begin{aligned}
\mathbf{I} = \mathbf{\{}\{X_1,X_2,...\}, O_{first},Y,Z \mathbf{\}}.
\label{eq:data_format2}
\end{aligned}
\end{equation}

Practically, in this paper, we constitute future observations based on multi-frame, multi-target action descriptions combined with human priors, and further process them with GPT-4~\cite{GPT4} to ultimately form reasonable future inferences. More details are provided in the supplementary materials. 

Figure~\ref{fig.3} provides an illustrative example of FIT, when a user questions Merlin about the future of a player in red attire, Merlin initially presents the observed trajectory of the concerned player, followed by the trajectory of another player in white. Using these trajectories, Merlin deduces that \textit{the player in white is likely to tackle the one in red, resulting in both players falling to the ground.}

\noindent\textbf{Training Details.} 
We freeze the vision encoder and keep the convolutional projector and the LLM unfreezed in this stage. On this basis, we primarily adopt the open-source instruction tuning datasets, \ie, LLaVA-665K~\cite{llava1.5}, for building the basic ability for multi-round visual-language conversation. For further unleash the foresight minds of model, we first uniformly sample a certain number of multitask dialogues in FPT, in order to maintain the model's capacity of modeling the dynamic clues across multi-frame images. In addiction, we also sample data from three specific scenario datasets~\cite{li2021multisports,malla2020titan,wu2021star} and construct around 60K FIT conversations based on the aforementioned data construction process.

\section{Experiment}
\label{exps}

\subsection{Experimental Settings}
\label{exp_setting}

\noindent \textbf{Datasets.} For the foresight pre-training (FPT) stage, we first use 10M image-text pairs sampled from LAION400M~\cite{schuhmann2021laion} to ensure multimodal alignment. On this basis, we gather various open-source datasets with rich annotations to conduct multi-task learning, including (1) \textbf{object detection} datasets: Object365~\cite{shao2019objects365} and OpenImage~\cite{kuznetsova2020open}; (2) \textbf{tracking} datasets: LaSOT~\cite{fan2019lasot}, GOT10K~\cite{huang2019got}, MOT17~\cite{milan2016mot16}, DanceTrack~\cite{sun2022dancetrack} and SOMPT22~\cite{simsek2022sompt22}; (3) \textbf{grounding} dataset: Ref-COCO~\cite{refcoco}; (4) \textbf{object relation} dataset: VCR~\cite{vcr}.
For these data, as described in Section~\ref{FPT}), we apply strict task definitions and format specifications, and re-organize them in the form of interleaved frames and observations. Ultimately, we obtain approximately 5M question-answer data, which are mixed with 10M paired data for foresight pre-training.

For the foresight instruction-tuning (FIT) stage, we mix approximately 730K conversation data, including (1) open-source instruction-tuning data LLaVA-665K~\cite{llava1.5}, which integrates a series of VQA datasets~\cite{TextVQA} and multi-round conversation datasets~\cite{llava}; (2) around 30K FIT multi-frame conversations constructed from three specific scenarios including MultiSports~\cite{li2021multisports}, TITAN~\cite{malla2020titan} and STAR~\cite{wu2021star} based on the data construction method described in Section~\ref{FIT}; (3) nearly 40K randomly sampled FPT multi-task data. For more details of the datasets, please refer to the supplementary materials.

\input{Tabs/future_reasoning}

\noindent \textbf{Implementation Details.} 
As outlined in Section~\ref{overall}, Merlin utilizes the CLIP-ViT-L/14~\cite{clip} as its vision encoder for image encoding and the open-source Vicuna-7B v1.5~\cite{vicuna} for foresight decoding. Between them, a $3 \times 3$ convolution layer with padding set to $1$ and a stride of $2$ is employed for both dimension projection and token aggregating.
During the foresight pre-training, we optimize all parameters of the model, setting the learning rate to $5e-5$ and training for one epoch. In the instruction tuning stage, we freeze the visual encoder and fine-tune the parameters of the projector and LLM. In both stages, we train Merlin using the AdamW~\cite{AdamW} optimizer and a cosine annealing scheduler~\cite{loshchilov2016sgdr} as the learning rate scheduler. 
The entire training process is conducted on 64 NVIDIA A800 GPUs, with approximately 12 hours required for pre-training and 3 hours for instruction-tuning. 
Additional implementation details can be found in the supplementary materials.

\subsection{Properties Evaluation of Foresight Minds}
\label{future exp}

In this section, we mainly verify the foresight minds (future reasoning) of Merlin from two aspects, \ie, prediction reasoning and identity association ability, where the former focuses on forecasting and reasoning location, events or behavior based on image observation, and the latter focuses on the model's ability to establish subject identity associations across multiple frames to obtain dynamic clues for future reasoning.

\noindent \textbf{Prediction Reasoning.} To evaluate this ability, we probe this ability based on the several sub-tasks of MMBench~\cite{liu2023mmbench}. MMBench provides a comprehensive evaluation system to assess various capabilities of MLLM, with some metrics focusing on the model's prediction and reasoning capabilities. To this end, we pick out these metrics to establish this new future reasoning benchmark and compare Merlin with the existing SOTA models. As shown in Table~\ref{tab:feature_reason}, Merlin achieves the best overall performance ($64.4$ average score on the development set and $66.5$ average score on the test set). Moreover, it obtains the best in $8/10$ indicators and ranks second in all other indicators, which favorably demonstrates Merlin's strong predcition and reasoning ability.

\noindent \textbf{Identity Association.} 
We examine this ability by evaluating the performance of object-tracking tasks, which can comprehensively demonstrate object association and prediction capabilities.
To this end, we evaluate Merlin in existing mainstream tracking benchmarks, \ie, LaSOT~\cite{fan2019lasot} and GOT10K~\cite{huang2019got}. It is worth noting that Merlin is the \textbf{\emph{first} MLLM that can also carry out tracking tasks}. As shown in Table~\ref{tab:track}, Merlin achieves comparable performance with expert models and even outperforms on some metrics. Notably, we only \textit{sample a small amount} of tracking data to train Merlin instead of the full amount of data, which means LLM exhibits significant potential in handling temporal tasks, possibly because tracking, as a temporal task, can be viewed as a casually frame-level autoregressive task.

\begin{table}[t]
  \centering

\begin{minipage}[t]{0.48\textwidth}
\makeatletter\def\@captype{table}
\input{Tabs/future}
\end{minipage}
\begin{minipage}[t]{0.49\textwidth}
\makeatletter\def\@captype{table}
\input{Tabs/mmbench}
\end{minipage}
\vspace{-5mm}
\end{table}

\subsection{General Comprehension}
\label{exp_task}

In order to showcase the general multi-modal ability, we further benchmark Merlin on various VQA benchmarks and recent benchmarks proposed for evaluating the comprehensive capabilities of MLLMs.

\noindent \textbf{Visual Question Answering (VQA).} We first evaluate Merlin on several mainstream VQA benchmarks to reflect the perceptual abilities of MLLMs in understanding image content. As shown in Table~\ref{tab:general}, Merlin achieves competitive performance compared with existing advanced MLLMs in the selected VQA benchmarks (VQA). The results indicate that Merlin possesses strong image understanding and question-answering capabilities. 

\noindent \textbf{Synthetica MLLM Benchmarks.} 
Recently, several benchmarks have been proposed to evaluate the comprehensive performance of MLLMs, encompassing diverse finer-grained scenarios including visual perception, object recognition, optical character recognition (OCR), future reasoning, and so on. In this part, we select several mainstream MLLM benchmarks to evaluate Merlin. As shown in Table~\ref{tab:general}, We present performance in accuracy on benchmarks including MM-Vet~\cite{yu2023mmvet} and MMBench~\cite{liu2023mmbench}. On MMBench, we report results on the both development and test sets. 
The results show that Merlin significantly outperforms comparative methods, even though many methods utilized a substantial amount of in-house data for pre-training, or employed several times more parameters. This implies that, while introducing foresight minds into MLLMs, we not only preserved their original visual capabilities but even \textit{further enhanced their overall level of visual perception.}

\input{Tabs/pope}

\subsection{Object Hallucination}
\label{exp_pope}

Hallucination presents a significant challenge in existing MLLMs. This term describes the phenomenon where the generated textual content exhibits inconsistencies when compared to its corresponding image content. In this section, we present the experiments from the Polling-Based Object Probing Evaluation (POPE~\cite{li2023pope}). As demonstrated in Table~\ref{tab:pope}, Merlin surpasses recent SOTA methods with clear margins. More specifically, Merlin achieves optimal performance in all metrics across three scenarios: \textbf{Random}, \textbf{Popular} and \textbf{Adversarial}, with improvements of up to $\bf{5}$ points compared to the highly competitive baseline Shikra~\cite{shikra}. Surprisingly, in multiple scenarios, the 'yes' rate of Merlin is quietly close to $50\%$, demonstrating its extraordinary visual perception capabilities.

We analyze this success largely owing to the proposed foresight learning (FPT and FIT). By enabling the model to learn the dynamic correspondence between trajectories across multiple images, the model has gained a \textit{more precise ability to attend to relevant object (trajectories) contexts in the image}, which helps to better avoid misidentification and misalignment of irrelevant targets. We believe that this result will provide new thinking about addressing the issue of hallucinations in MLLM.

\input{Tabs/ablation_merlin_v1}

\input{Tabs/ablation_res_proj}

\subsection{Ablative Analysis of FPT \& FIT}
\label{exp_discuss}

As introduced in Section~\ref{FPT} and Section~\ref{FIT}, FPT serves as the pre-training strategy to enable MLLM to encapsulate dynamic information across frames by predicting the trajectory of the next frame. FIT is designed to activate the ability of foresight minds in a way of \textbf{\textit{Trajectory CoT}} during instruction fine-tuning. To further explore the effect of FPT and FIT, we conduct an ablation study based on the established future reasoning benchmark and tracking dataset GOT10K~\cite{huang2019got}. As shown in Table~\ref{tab:strategies}, we mainly report the average overlap (AO) of GOT10K and the average score of future reasoning in the development set.

The results show that both FPT and FIT training strategies contribute to the improvement of the metrics. Combining both FPT and FIT, Merlin achieves the best performance which proves the effectiveness of the proposed strategies. Furthermore, we can also observe that the lack of image-text pair data during the pre-training stage considerably hampers the model's general ability. This phenomenon supports our perspective that, during the comprehensive pre-training phase, the integration of image-text pair data is essential for maintaining modality alignment and preventing a decline in combined capabilities.

\subsection{Ablative Analysis of Model Configuration}
\label{exp_ablation}

The configuration of model architecture for large-scale models is also a focal point of interest for researchers. In this subsection, we specifically investigate the impact of Merlin's model configuration on performance. As depicted in Table~\ref{tab:ablation_res_proj}, we focus on examining the effects of model input resolution, the visual encoder of the model, and the model's projector on the ultimate performance of Merlin. From the experimental outcomes, we can draw the following conclusions:

\noindent \textbf{(i)} High-resolution input is more conducive to visual perception and understanding tasks (row \ding{202} and \ding{203}), particularly for tasks that require precise localization, such as detection and tracking.
 
\noindent \textbf{(ii)} The primary contribution of Conv2d is the ability to compress the number of tokens efficiently and elegantly, which is crucial for supporting high-resolution images. In contrast, MLPs cannot compress tokens. This high token count hinders the training with multiple images. Moreover, more visual tokens does not improve performance in future reasoning tasks (row \ding{202} and \ding{204}). We speculate that an increased number of visual tokens may lead to the sparsity of supervision.

\noindent \textbf{(iii)} During the pre-training phase, the visual encoder should be unfrozen (row \ding{202} and \ding{205}), which is beneficial for modal alignment and the expansion of the fine-grained spatial information. Similar conclusion is also claimed in \cite{chen2024spatialvlm}.

\begin{figure*}[t]
\centering
\includegraphics[width=1.0\linewidth]{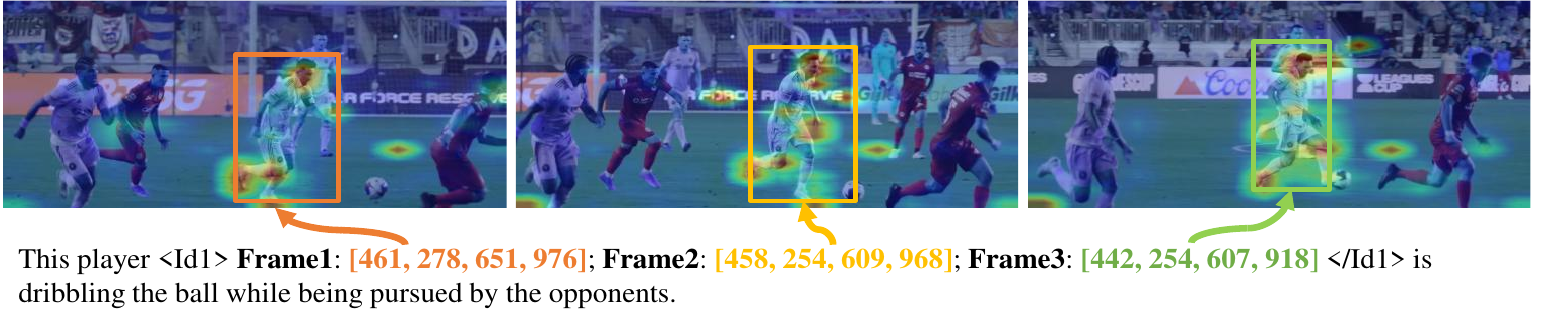}
\caption{\textbf{Attention map visualization.} To facilitate the observation, we map the attention between the box responses and the visual tokens of each frame for visualization.}
\vspace{-2mm}
\label{fig:attn}
\end{figure*}

\subsection{Visualization Analysis}
\label{exp_vis}

In this subsection, we visualize the attention map of Merlin to further substantiate the effectiveness of utilizing the proposed strategies. As shown in Figure~\ref{fig:attn}, we select the output attention map of the middle-level layers of LLM for visualization. We can observe that the word embedding of the output trajectory coordinates can attend to the corresponding object from different frames correctly. This visualization results further prove that the trajectory representation is a good interface to enable MLLM to establish the alignment between the language description and the multi-images dynamic visual contexts. 
Furthermore, this effectively explains why Merlin possesses a more powerful comprehensive visual capability and a greatly lower level of hallucination compared to previous baselines. Indeed, \textit{the trajectory-driven foresight learning allows the large language model to \textbf{read} images more profoundly!}

\section{Limitation and Conclusion}
\label{conclusion}

This study highlighted an obvious deficiency in Multimodal Large-Language Models (MLLMs), specifically their ability to predict future events or outcomes based on current observations, referred as \textit{“foresight minds”}. To address this, we serve as the first to point out that trajectory, as a highly structured representation, is a good learning objective to assist MLLM in obtaining dynamic information from the image observations. Based on this insight, we introduced a unique training method including \textit{Foresight Pre-Training (FPT)} and \textit{Foresight Instruction-Tuning (FIT)}. By synergizing FPT and FIT, we created \textbf{\textit{Merlin}}, a unified MLLM that effectively understands and outputs spatial coordinates or tracklets from single images or multiple frames. Merlin excels at a range of traditional vision-language tasks while demonstrating powerful future reasoning capacities. Despite the substantial advancements made by Merlin, there still are some limitations, particularly in processing long sequential videos and more comprehensive future reasoning evaluation. We aspire for Merlin to guide the enhancement of more advanced MLLMs in the future.

\section*{Acknowledgements}

This work was supported by the National Natural Science Foundation of China under Grant 62176096 and Grant 61991412.

\appendix

\section{Appendix}
\label{appdix}

In this supplementary material, we offer additional information about \textbf{Merlin} due to the paper's page limit of pages. Specifically, Section~\ref{dataset} provides in-depth insights into the dataset we constructed, including its data sources and how it was created. Section~\ref{format} delves into the data formats of the proposed FPT and FIT, as explained in the main manuscript. Section~\ref{training} offers a more detailed explanation of the training approach. Section~\ref{dis} expands on the discussion of the proposed method, including its limitations and future directions. Finally, Section~\ref{exp} offers additional experimental results and demo visualizations.

\section{Dataset Details}
\label{dataset}

In Section 3 of the manuscript, we explained how we created the Foresight Pre-Training (FPT) and Foresight Instruction-Tuning (FIT) datasets. Now, in this section, we go into greater detail about how we collected and built the dataset. To start, we provide an overview of our collected data in Table~\ref{tab:all_data}, and then dive into the step-by-step process of how it was constructed.

\input{Tabs/tab1}

\begin{figure*}[t]
\centering
\includegraphics[width=1.0\linewidth]{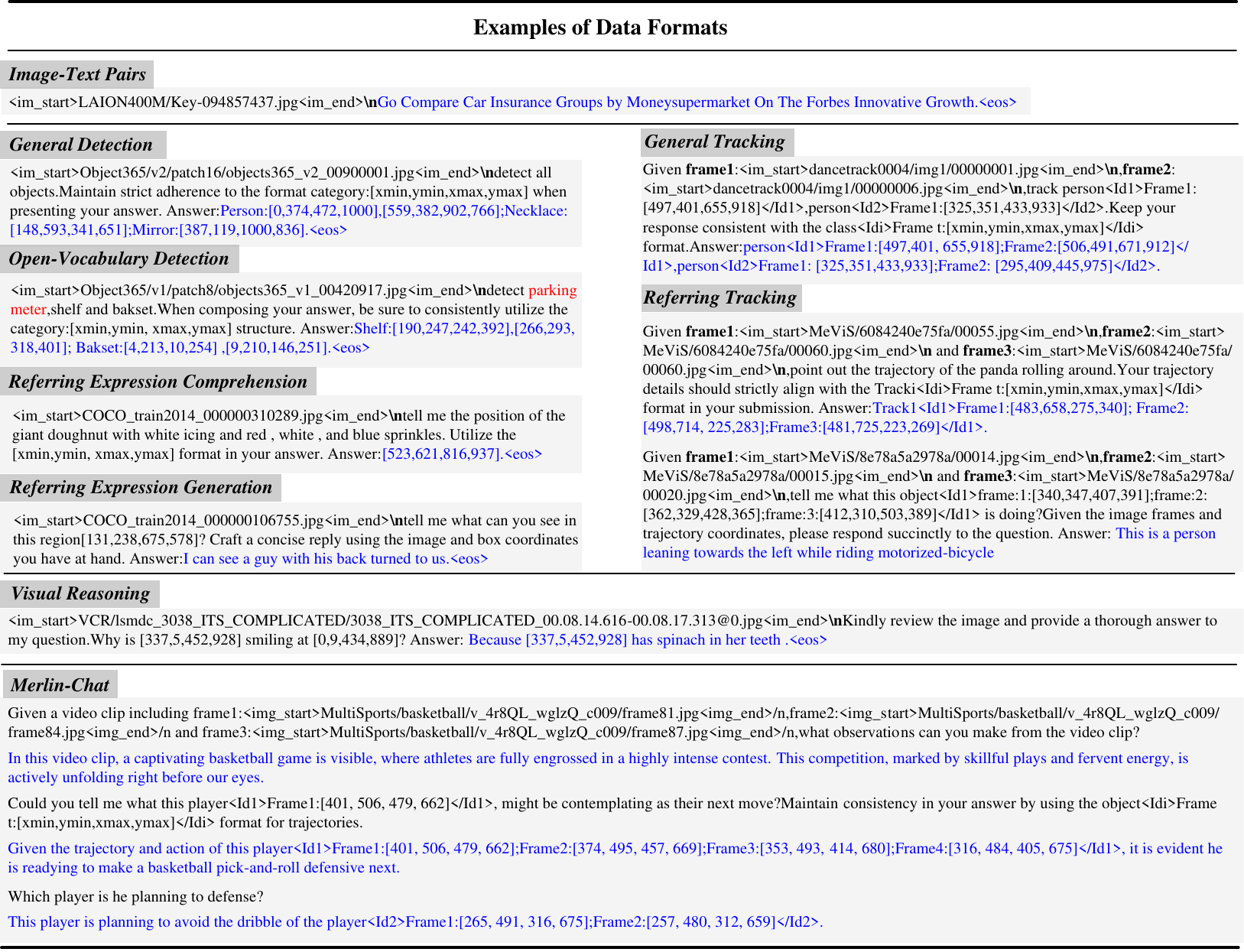}
\caption{\textbf{Data format visualization}. In our training, we use datasets that involve multiple tasks. To illustrate how these datasets are structured, we've chosen an example from each one. It's important to mention that all information about boxes has been adjusted to a standard range of 1000. In the examples, questions are shown in black text, answers in \textcolor{blue}{blue}, and negative samples in \textcolor{red}{red}.}
\label{fig:format1}
\end{figure*}

\noindent \textbf{Image-text pairs.} We mainly collect image-text pairs dataset from the open-sourced dataset, LAION-400M~\cite{schuhmann2021laion}. LAION-400M is a high-quality dataset with CLIP-filtered 400 million image-text pairs. In this paper, we random sample $10$M image-text pairs of LAION-400M for the Foresight Pre-Training.

\noindent \textbf{Detection datasets.} Detection datasets are crucial for improving the model's ability to understand space during its initial training phase. In our research, we used datasets from various publicly available sources like Object365~\cite{objects365}, OpenImage~\cite{kuznetsova2020open}, and CrowdHuman~\cite{shao2018crowdhuman}. We processed these datasets in the following ways:

\noindent (1) Extract all objects in each image along with their categories and bounding boxes coordinates.

\noindent (2) Remove too small objects (smaller than $1/32$ of the image size).

\noindent (3) Randomly select fixed number (n = 15) of categories if image has more categories than a set limit.

\noindent (4) Compose the detection data format as the following:

\texttt{\textit{cat1:<box>,<box>,<box>;cat2:<box>;...}}.

As shown in Table~\ref{tab:all_data}, we also created special output prompts that guide the model to respond in a precise format, as detailed in our study.

\noindent \textbf{Tracking datasets.} Building tracking data is a key part of our Foresight Pre-Training (FPT) method, as we've discussed in our main paper. FPT is designed to causally modeling the trajectory in conjunction with each frame of the image to empower the model to acquire foresight minds. The tracking data naturally includes trajectory information about movement paths, making it ideal for our FPT task. We gathered a variety of open-source tracking data, including Single Object Tracking (SOT) and Multi-Object Tracking (MOT) datasets like GOT10K~\cite{huang2019got}, LaSOT~\cite{fan2019lasot}, MOT17~\cite{milan2016mot16}, Sompt~\cite{simsek2022sompt22}, DanceTrack~\cite{sun2022dancetrack}, SportsMOT~\cite{cui2023sportsmot}, BDD100K~\cite{yu2020bdd100k}, MeViS~\cite{ding2023mevis}), etc. We follow the steps below to pre-process the tracking data.

\noindent (1) Random sample 3 to 5 frames at a certain frame interval (gap = 3,4 or 5) to form a video clip each time.

\noindent (2) Extract all trajectories with their category, identity and bounding boxes in each video clip.

\noindent (3) Remove the trajectory containing too small objects (smaller than $1/32$ of the image size).

\noindent (4) Select the initial observation (location, appearance or action in the first frame) as the trajectory query.

\noindent (5) Compose the tracking data format as the following: 

\texttt{\textit{query,cat1<Idi>Frame1:<box>;Frame2:<box>;...</Idi>}}.

Similar to constructing detection data, we also adopt output format prompts to guide the model to provide answers in the desired trajectory format.

\noindent \textbf{Visual Reasoning datasets.} Visual reasoning dataset is constructed to enhance the common sense reasoning ability of the model. In this work, we mainly collect the VCR dataset and adopt the same processing method as Shikra~\cite{shikra}. 

\noindent \textbf{Referring datasets.} For referring dataset, we mainly collect from RefCOCO~\cite{refcoco} and MeViS~\cite{ding2023mevis} to construct image referring and video referring datasets, respectively. For image referring, we simply extract the description and the corresponding bounding box from each image. For video referring, we primarily extract pairs of trajectories and trajectory descriptions from MeViS.

\noindent \textbf{Dialogue datasets.} To facilitate the model's ability to achieve long conversations and follow user instructions, we utilized the open-sourced LLaVA-665K instruction tuning dataset~\cite{llava1.5}. Additionally, we created a smaller instruction tuning dataset (30K entries) using our \textit{Trajectory Chain-of-Thought (T-CoT)} method within Foresight Instruction Tuning. This dataset focuses on three specific contexts: MultiSports~\cite{li2021multisports}, TITAN~\cite{malla2020titan}, and STAR~\cite{wu2021star}. MultiSports catalogs multi-person sports actions with spatial and temporal details. TITAN encompasses street scenes with varied labels like vehicle states, pedestrian age groups, and specific pedestrian actions. STAR offers real-world video data with symbolic descriptions and logic-based analysis.

To create the T-CoT conversation data, we leveraged GPT-4 to generate rich feature reasoning dialogues using the trajectories and actions from these datasets. We guided GPT-4 to interpret video clips through these trajectories and their associated descriptions or actions, then instructed it to formulate Q\&A pairs. These questions were designed to derive answers from available information. This process resulted in 30K QA pairs, incorporating trajectory coordinates in both questions and answers. We plan to expand this dataset, which we refer to as Merlin-chat, in the future.

\begin{table}[t]
    \centering
    \caption{\textbf{Training hyperparameters of Merlin}. The hyperparameter placed in the middle indicates that this hyperparameter is used in both stages.}
    \tablestyle{1pt}{1.0}
    \begin{tabular}{l cc}
         \toprule
         \textbf{Configuration}            & \textbf{Pre-training} & \textbf{Supervised Fine-tuning} \\
         \midrule
         ViT init.                & OpenAI-CLIP-L/14 & Merlin FPT \\
         LLM init.                & Vicuna-7B-v1.5 & Merlin FPT \\
         Projection init.         & random & Merlin FPT \\
         Image resolution         & $448^2$ & $448^2$ \\
         ViT sequence length      & 2048 & 2048 \\
         LLM sequence length      & 2048 & 2048\\
         Optimizer                & \multicolumn{2}{c}{AdamW} \\
         Optimizer hyperparameter & \multicolumn{2}{c}{$\beta_{2}=0.95, eps=1e^{-8}$} \\
         Peak learning rate       & \multicolumn{2}{c}{$5e^{-5}$} \\
         Minimum learning rate    & \multicolumn{2}{c}{0} \\
         ViT learning rate decay  & 0.9 & 0 \\
         ViT Drop path rate       & \multicolumn{2}{c}{0} \\
         Learning rate schedule   & \multicolumn{2}{c}{cosine decay} \\
         Weight decay             & \multicolumn{2}{c}{0.05} \\
         Gradient clip            & \multicolumn{2}{c}{1.0} \\
         Training steps           & 7k & 3k \\
         Warm-up steps            & 70 & 90 \\
         Global batch size        & 2048 & 256 \\
         Gradient Acc.            & 8 & 1 \\
         Numerical precision      & \multicolumn{2}{c}{$\mathtt{bfloat16}$} \\
         Optimizer sharding       & \multicolumn{2}{c}{\ding{51}} \\
         Activation checkpointing & \multicolumn{2}{c}{\ding{55}} \\
         Model parallelism        & \multicolumn{2}{c}{\ding{55}} \\
         Pipeline parallelism     & \multicolumn{2}{c}{\ding{55}} \\
         \bottomrule
    \end{tabular}
    \label{tab:hyperparam}
\end{table}

\section{Data Format Details}
\label{format}

To assist readers in comprehending the specific nature of the data we've constructed, this section includes visualizations of the training data format. As illustrated in Figure~\ref{fig:format1}, we present an exhaustive list of all data formats used, encompassing both pretraining and instruction tuning phases. Notably, we have incorporated negative samples (highlighted in bold red) into our question-answer pairs. This addition is designed to teach the model the skill of responding with a negative answer when appropriate, thereby bolstering its ability to resist producing hallucinatory responses.

\input{Tabs/ablation_task_prompt}
\input{Tabs/ablation_tcot}

\section{Training Details}
\label{training}

In this section, we report the detailed training hyperparameter settings of Merlin in Table~\ref{tab:hyperparam}.

\section{More Discussions}
\label{dis}


\noindent \textbf{Limitations and Future Work.} While Merlin demonstrates exceptional foresight capabilities, it is not without its limitations in its current form. A significant constraint is the inability to support long-range video sequences (exceeding 8 frames), which hinders its capacity to model extended motion information. This is primarily due to Merlin's reliance on an image encoder rather than a video encoder, resulting in an excessive number of vision tokens for the LLM to process. Addressing this, the development of a more efficient tokenizer for long-range videos emerges as a crucial area for future research. Moreover, this paper introduces a novel benchmark for future reasoning, building upon the existing MMBench framework. Currently, there is no comprehensive and accurate benchmark to evaluate future reasoning abilities thoroughly. Therefore, exploring the creation of a more robust and comprehensive future reasoning benchmark represents another significant avenue for future investigation.

\begin{figure*}[t!]
\centering
\includegraphics[width=1.0\linewidth]{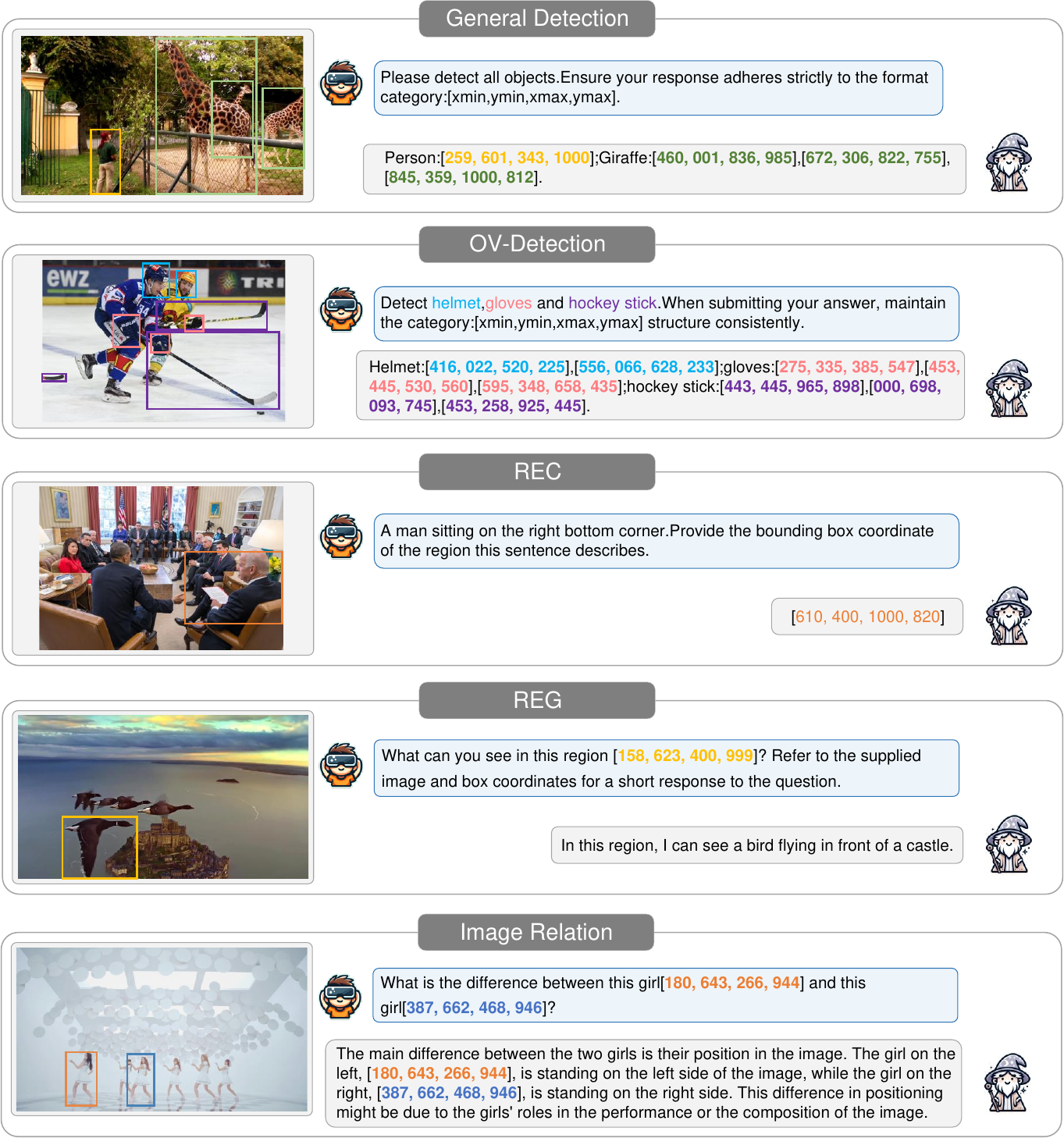}
\caption{\textbf{More conversation visualization using Merlin in image-level tasks}. We showcase additional examples where Merlin adeptly handles various image-level tasks including general detection, open vocabulary detection (OV-Detection), referring expression comprehension (REC), referring expression generation (REG), and relation reasoning.}
\label{fig:image_cases}
\vspace{3cm}
\end{figure*}

\begin{figure*}[t!]
\centering
\includegraphics[width=1.0\linewidth]{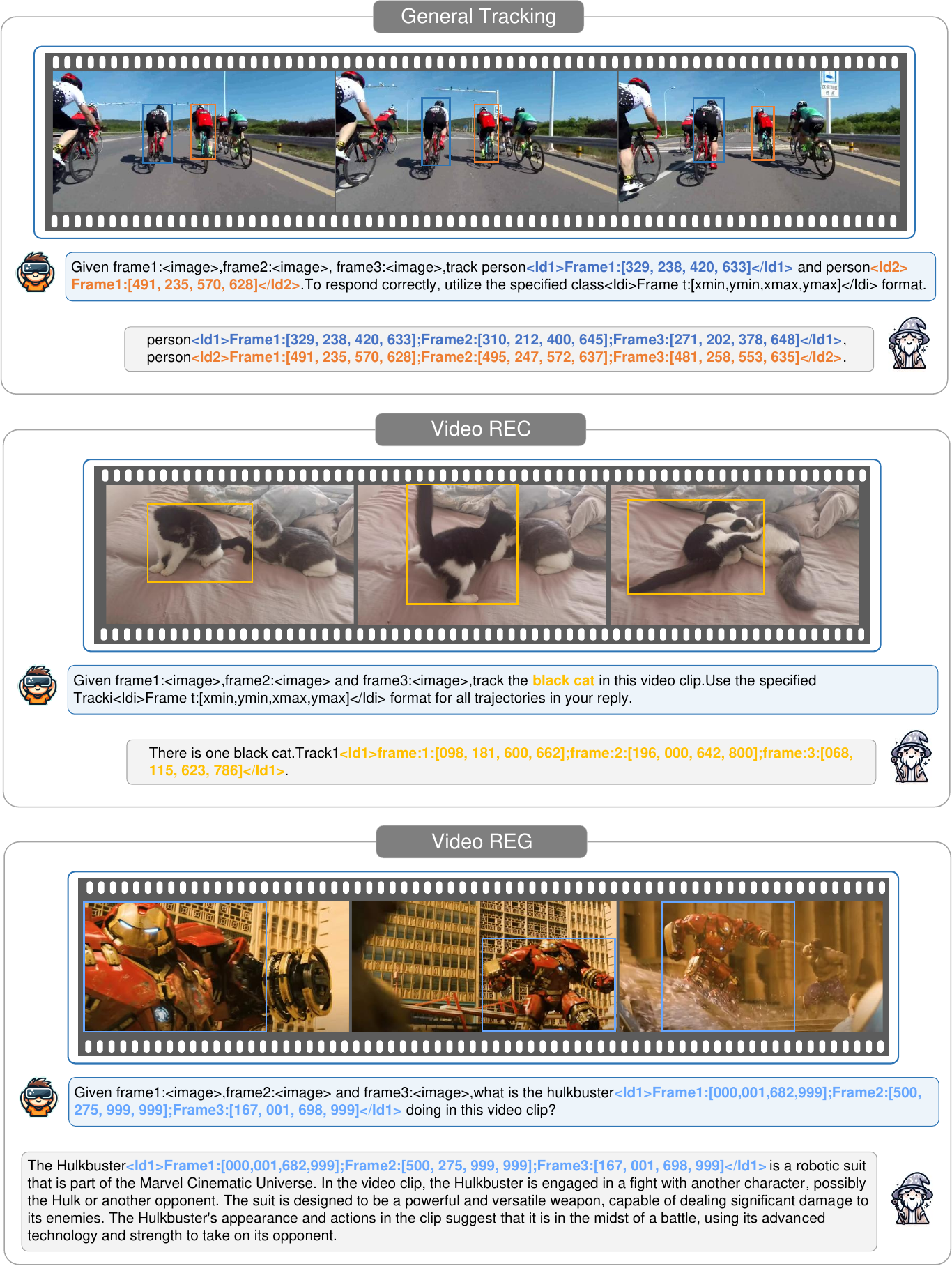}
\caption{\textbf{More conversation visualization using Merlin in video-level tasks}. We showcase additional examples where Merlin adeptly handles various video-level tasks including general tracking, video referring expression comprehension (Video REC) and video referring expression generation (Video REG).}
\label{fig:video_cases}
\end{figure*}

\begin{figure*}[t!]
\centering
\includegraphics[width=1.0\linewidth]{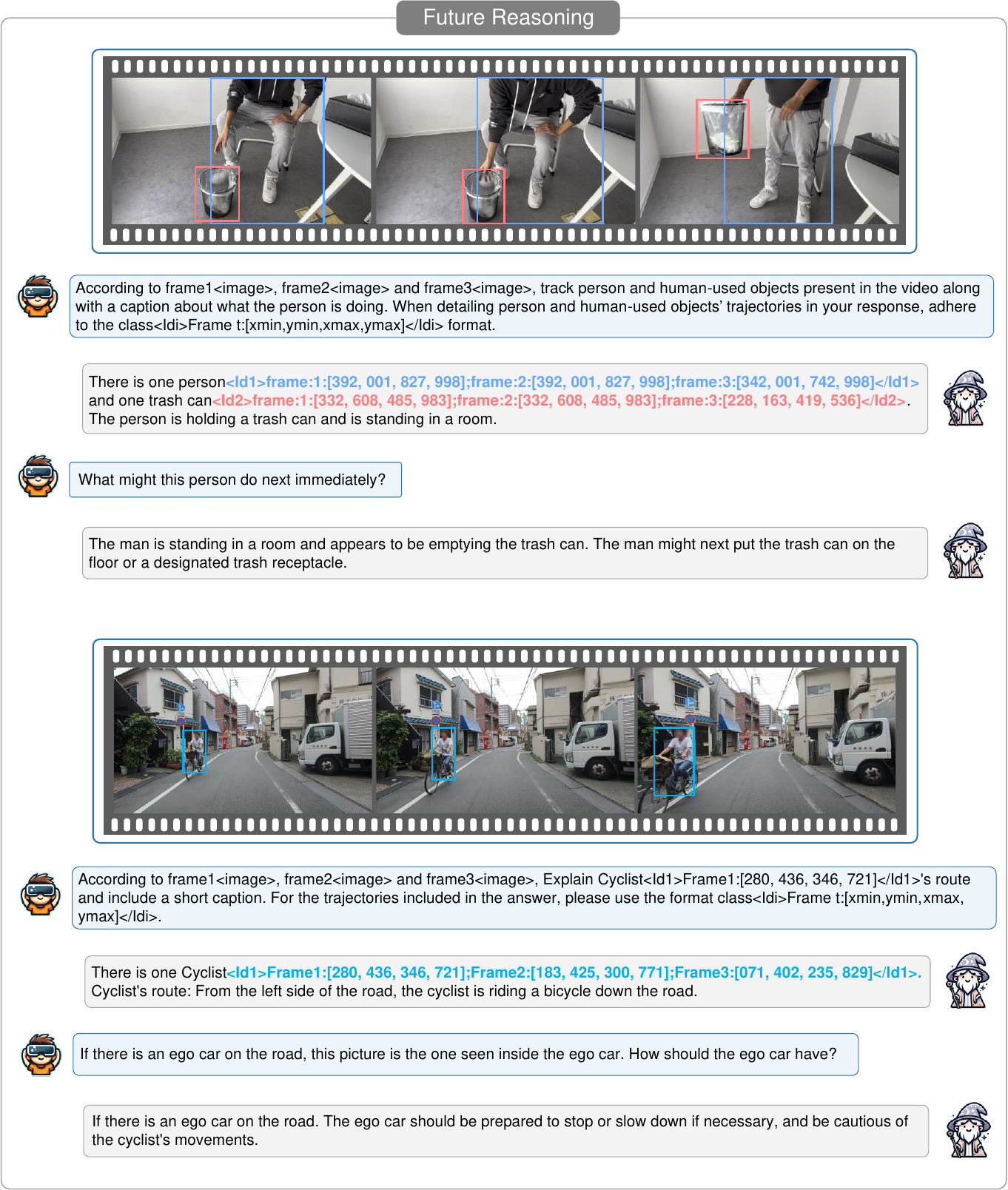}
\caption{\textbf{More future reasoning conversation visualization using Merlin}.}
\label{fig:future_reason_cases}
\end{figure*}

\section{More Experiments}
\label{exp}

In this section, we aim to provide more experimental results including ablation studies (shown in Table~\ref{tab:task} and Table~\ref{table:tcot}) and qualitative analysis (shown in Figure~\ref{fig:image_cases},~\ref{fig:video_cases} and ~\ref{fig:future_reason_cases}).

\noindent \textbf{Effect of Precise Task Description.} In the method section of main paper, we emphasize the importance of precise task descriptions, which can prevent conflicts between multiple task learning and damage to general language abilities. The performance corresponding to the models with and without the precise task description is reported in Table~\ref{tab:task}. We can observe that the model without the precise task description behaves significantly poorer, especically the tracking performance ($51.4\%$ - $28.4\%$). The results further prove that precise task description is an important option for multi-task learning in existing MLLM.

\noindent \textbf{The effectiveness of trajectory CoT (T-CoT).} As shown in Tab.~\ref{table:tcot}, despite the fact that FPT has endowed the model with the capability for future reasoning, the actual performance of prediction reasoning remains limited. However, with the incorporation of the T-CoT, there has been a significant enhancement in the model's reasoning abilities. This aligns with the mainstream viewpoint on the CoT, which posits that the model's capabilities need to be progressively guided through a step-by-step reasoning process. Furthermore, T-CoT enables the model to visually analyze the image by forecasting the trajectory, which provides richer dynamic clues for the following reasoning. 

%
%
\bibliographystyle{splncs04}
\bibliography{main}
\end{document}

%% file: Tabs/future_reasoning.tex
\begin{table*}[!t]
        \centering
        \large
        \caption{\textbf{The Effectiveness of Prediction Reasoning.} We mainly select 5 metrics from MMBench develop and test set, respectively, including \textbf{OL}: Object localization (Prediction), \textbf{PPR}: Physical property
reasoning, \textbf{FR}: Function reasoning, \textbf{IR}: Identity reasoning, and \textbf{FP}: Future prediction. \textbf{Avg.} denotes the average score. The best and second-best performances are shown in bold font and underlined respectively.
}
\resizebox{1.0\columnwidth}{!}{
	\begin{tabular}{l c c p{8mm}p{9.5mm}p{8mm}p{8mm}p{8mm} c c p{8mm}p{9.5mm}p{8mm}p{8mm}p{8mm}}
\toprule[.9pt]
    \multirow{2}{*}{\textbf{Method}} & \multirow{2}{*}{\textbf{LLM Size}} & \multicolumn{6}{c}{Prediction Reasoning (Dev.)} && \multicolumn{6}{c}{Prediction Reasoning (Test)} \\
    \cmidrule{3-8}
    \cmidrule{10-15}
             & & \textbf{Avg.} & \textbf{OL} & \textbf{PPR} & \textbf{FR} & \textbf{IR} & \textbf{FP} && \textbf{Avg.} & \textbf{OL} & \textbf{PPR} & \textbf{FR} & \textbf{IR} & \textbf{FP} \\
            \midrule
            InstructBLIP~\cite{dai2023instructblip} & 13B & 42.0 & 14.8 & 30.7 & 56.8 & 88.9 & 19.0 && 44.4 & 5.7 & 24.0 & 67.3 & \underline{92.7} & 32.4 \\
            MiniGPT-4~\cite{minigpt4}  & 13B & 43.3 & 28.4 & 30.7 & 49.4 & 86.7 & 21.4 && 48.9 & 21.0 & 35.0 & 67.3 & 90.2 & 31.1 \\
		OpenFlamingo~\cite{openflamingo}  & 7B            & 5.28 & 2.5 & 10.7 & 8.6 & 2.2 & 2.4 && 11.5 & 2.9 & 14.0 & 9.3 & 11.0 & 20.3\\
            MMGPT~\cite{liu2023mmbench}        & 7B            & 19.5  & 1.2 & 24.0 & 9.9 & 60.0 & 2.4  && 16.8  & 3.8 & 13.0 & 12.1 & 52.4 & 2.7 \\
            MiniGPT-4~\cite{minigpt4}  & 7B       & 26.8 & 7.4 & 14.7 & 19.8 & 80.0 & 11.9  && 27.9  & 8.6 & 13.0 & 29.9 & 61.0 & 27.0\\
            InstructBLIP~\cite{dai2023instructblip} & 7B           & 34.8 & 6.2 & 17.3 & 51.9 & 84.4 & 14.3  &&  39.0 & 2.9 & 17.0 & 52.3 & 78.0 & 44.6\\
            LLaVA~\cite{llava} & 7B           & 38.7 & 8.6 & 25.3 & 53.1 & 77.8 & 28.6  && 39.7 & 13.3  & 35.0 & 48.6 & 82.9 & 18.9\\
            mPLUG-Owl~\cite{ye2023mplug}  & 7B          & 41.0 & 18.5 & 18.7 & 66.7 & 86.7 & 14.3 &&  45.9 & 16.2 & 23.0 & 59.8 & 91.5 & 39.2\\
            Shikra~\cite{shikra}  & 7B          & 51.5 & 32.1 & 30.7 & 63.0 & 88.9 & 42.9  && \underline{60.0}  & 27.6 & \underline{50.0} & \underline{70.1} & \underline{92.7} & 59.5\\
            Kosmos-2~\cite{KOSMOS} & 1.6B           & 54.4 & 38.3 & 33.3 & 56.8 & 91.1 & \underline{52.4} &&  58.2 & \underline{40.4} & 30.0 & 65.4 & 89.0 & \bf66.2\\
            LLaVA-1.5~\cite{llava1.5} & 7B       &   
            \underline{59.6} & \bf43.2 & \underline{52.0} & \underline{71.6} & \underline{93.3} & 38.1 && - & - & - & - & - & - \\
    \midrule    
    \rowcolor{aliceblue!60} \textbf{Merlin} (\textbf{Ours}) & 7B   & \bf64.4 & \underline{42.0}	& \bf54.7	& \bf72.8 & \bf97.8 & \bf54.8 && \bf66.5 & \bf41.3 & \bf51.0 & \bf83.0 & \bf97.6 & \underline{59.7}\\
  \bottomrule[.9pt]
	\end{tabular}}
        \setlength{\abovecaptionskip}{0.2cm}
\label{tab:feature_reason}
\end{table*}

%% file: Tabs/future.tex
        \large
        \caption{\textbf{Comparison on main tracking benchmarks.} Notably, the original LLaVA-1.5~\cite{llava1.5} model was incapable of performing tracking tasks. Therefore, we utilized the model configuration of LLaVA-1.5 and trained a version of the model with the same dataset as Merlin for the fair comparsion.}
\resizebox{1.0\columnwidth}{!}{
	\begin{tabular}{lcccccccc}
  \toprule[.9pt]
  & \multicolumn{3}{c}{LaSOT} &&\multicolumn{3}{c}{GOT10k} \\
    \cmidrule{2-4}  \cmidrule{6-8}
            \textbf{Method} & \textbf{Success} & \textbf{P}$_{norm}$ & \textbf{P} && \textbf{AO} & SR$_{0.5}$ & SR$_{0.75}$ \\
            \midrule
            \multicolumn{8}{l}{\small{\textbf{\emph{Specialist Models}}}} \\
		SiamFC~\cite{siamFC}              & 33.6 & 42.0 & 33.9 & & 34.8 & 35.3 & 9.8 \\
            ATOM~\cite{ATOM}                 &   51.5 & -  &- & & 55.6 & 63.4 & 40.2\\
            SiamRPN++~\cite{SiamRPN++}      &  49.6 & 56.9 & 49.1 & & 51.8 & 61.8 & 32.5\\
            SiamFC++~\cite{SiamFC++}          &  54.4 & 62.3 & 54.7 & & 59.5 & 69.5 & 47.9\\
    \midrule    
        \multicolumn{8}{l}{\small{\textbf{\emph{Generalist Models}}}} \\
        LLaVA-1.5~\cite{llava1.5}  & 19.4 & 16.5 & 12.8 & & 23.5 & 20.2 & 9.7 \\
    \rowcolor{aliceblue!60} \textbf{Merlin (Ours)} & 39.8   & 40.2  & 38.1 && 51.4	& 55.9	& 42.8\\
  \bottomrule[.9pt]
	\end{tabular}
        }
        \label{tab:track}

%% file: Tabs/mmbench.tex
        \large
        \caption{\textbf{Comparison with SOTA methods on main MLLM benchmarks.} For VQA tasks, we mainly choose GQA~\cite{hudson2019gqa} and VisWiz~\cite{gurari2018vizwiz} to evaluate the model; For general evaluation, we mainly choose MMBench~\cite{liu2023mmbench} and MM-Vet~\cite{yu2023mmvet}. $^\dagger$Includes using in-house data that is not publicly accessible.}
\resizebox{1.0\columnwidth}{!}{
	\begin{tabular}{lcccccc}
  \toprule[.9pt]
  & \multicolumn{2}{c}{VQA Task} &\multicolumn{3}{c}{Generalist} \\
    \cmidrule(rl){2-3}  \cmidrule(rl){4-6}
            \textbf{Method} & \textbf{GQA} & \textbf{VisWiz} & \textbf{MMB$_{d}$} & \textbf{MMB$_{t}$} & \textbf{MM-Vet}  \\
            \midrule
		BLIP-2~\cite{BLIP2}                     &  41.0 & 19.6 & -    & -    & 22.4 \\
            InstructBLIP~\cite{dai2023instructblip} &  49.2 & 34.5 & 36.0 & 33.9 & 26.2 \\
            Shikra~\cite{shikra}                    &     - &    - & 58.8 & 60.2 & -\\
            IDEFICS-9B~\cite{laurenccon2023obelics}                      &  38.4 & 35.5 & 48.2 & 45.3 & - \\
            IDEFICS-80B~\cite{laurenccon2023obelics}                     &  45.2 & 36.0 & 54.5 & 54.6 & -\\
            Qwen-VL$^\dagger$~\cite{qwen}                     &  59.3 & 35.2 & 38.2 & 32.2 & -\\
            Qwen-VL-Chat$^\dagger$~\cite{qwen}                &  57.5 & 38.9 & 60.6 & 61.8 & -\\
            LLaVA-1.5~\cite{llava1.5}               &  \bf62.0 & 50.0 & 64.3 & 59.5 & 30.5 \\
    \midrule    
    \rowcolor{aliceblue!60} \textbf{Merlin (Ours)}  &  60.5 & \bf50.4 &	\bf66.2 & \bf65.5 & \bf34.9 \\
  \bottomrule[.9pt]
	\end{tabular}
        }
        \label{tab:general}

%% file: Tabs/pope.tex
\begin{table*}[t]
\footnotesize
\centering
\large
\caption{\textbf{Zero-shot object hallucination evaluation on the COCO validation set.} ``Yes'' represents the proportion of positive answers that the model outputs.}
\resizebox{1.0\columnwidth}{!}
{
\begin{tabular}{lcccccccccc}
\toprule[.9pt]
\multirow{2}{*}{\textbf{Method}} & \multirow{2}{*}{\textbf{LLM Size}} & \multicolumn{3}{c}{\textbf{Random}} & \multicolumn{3}{c}{\textbf{Popular}} & \multicolumn{3}{c}{\textbf{Adversarial}} \\ 
\cmidrule(rl){3-5} \cmidrule(rl){6-8} \cmidrule(rl){9-11} & & Accuracy & F1-Score & Yes & Accuracy  & F1-Score & Yes & Accuracy & F1-Score  & Yes \\ \midrule
  LLaVA~\cite{llava} &  13B &  64.12 &  73.38 &  83.26 &  63.90 &  72.63 &  81.93 &  58.91 &  69.95 &  86.76 \\
  MiniGPT-4~\cite{minigpt4} &  13B &  79.67 &  80.17 &  52.53 &  69.73 &  73.02 &  62.20 &  65.17 &  70.42 &  67.77  \\ 
  InstructBLIP~\cite{dai2023instructblip} &  13B &  88.57 &  89.27 &  56.57 &  82.77 &  84.66 &  62.37 &  72.10 &  77.32 &  73.03  \\ 
  Shikra~\cite{shikra} &  13B &  86.90 &  86.19 &  43.26 &  83.97 &  83.16 &  45.23 &  83.10 &  82.49 &  46.50 \\
 MultiModal-GPT~\cite{gong2023multimodal} & 7B & 50.10 & 66.71 & 99.90 & 50.00 & 66.67 & 100.00  & 50.00 & 66.67 & 100.00 \\
 mPLUG-Owl~\cite{ye2023mplug} & 7B & 53.97 & 68.39 & 95.63 & 50.90 & 66.94 & 98.57 & 50.67 & 66.82 & 98.67 \\
 LLaVA~\cite{llava} & 7B & 72.16 & 78.22 & 76.29 & 61.37 & 71.52 & 85.63 & 58.67 & 70.12 & 88.33  \\ 
 LLaVA-1.5~\cite{llava1.5} & 7B & 83.29 & 81.33 & - & 81.88 & 80.06 & - & 78.96 & 77.57 & - \\
 Qwen-VL~\cite{qwen} & 7B & 84.73 & 82.67 & - & 84.13 & 82.06 & - & 82.26 & 80.37 & - \\ 
 \midrule
 \rowcolor{aliceblue!60} \textbf{Merlin (Ours)} & 7B & \textbf{91.58} & \textbf{91.66} & 49.38 & \textbf{89.53} & \textbf{89.56} & 50.27 & \textbf{84.10} & \textbf{84.95} & 55.63 \\ 
\bottomrule[.9pt]
\end{tabular}}
\label{tab:pope}
\end{table*}

%% file: Tabs/ablation_merlin_v1.tex
\begin{table}[t]
  \centering
\small
  \caption{\textbf{Ablation study of the proposed strategies in Merlin.} (ITP: Image-text pair data, ITD: instruction-tuning data). We mainly report the AO score of GOT10k and the average score of future reasoning.}
  \setlength{\tabcolsep}{3.0pt}
  \begin{tabular}{ccccc| ccccc}
    \toprule
\multicolumn{2}{c}{Pre-Training} && \multicolumn{2}{c}{Inst.-Tuning} &&   \multicolumn{1}{c}{GOT10K} && \multicolumn{2}{c}{Prediction Rea.} \\
\cmidrule{1-5}
\cmidrule{6-10}
ITP & FPT-Data && ITD & FIT-Data && \textbf{AO} && \textbf{Average$_{dev}$}\\
\midrule
\Checkmark & \xmark && \Checkmark & \xmark && - && 59.5\\
\Checkmark & \xmark && \Checkmark & \Checkmark && - && 60.7 \\
\xmark & \Checkmark && \Checkmark & \Checkmark && 15.5 && 52.8 \\
\Checkmark & \Checkmark && \Checkmark & \xmark && 51.4 && 61.2 \\
\rowcolor{aliceblue!60} \Checkmark & \Checkmark && \Checkmark & \Checkmark && \bf 51.4 && \bf 64.4 \\
    \bottomrule
  \end{tabular}
\label{tab:strategies}
\vspace{-2mm}
\end{table} 

%% file: Tabs/ablation_res_proj.tex
\begin{table*}[htbp]
  \centering
  \caption{\textbf{Ablation studies of the model settings} including resolution, vision encoder and projector of Merlin.}
  \resizebox{0.9\textwidth}{!}{
    \begin{tabular}{c|ccc|c|cc}
      \toprule[1pt]
       Exp &Resolution    &Projector & Visual Encoder & Tokens Num   &Prediction Rea.  & Got-10K\\ 
    \hline
        \ding{202} & 448x & Conv2d & unfrozen & 256 &  \bf 64.4 & \bf 51.4\\
        \ding{203} & 336x & Conv2d & unfrozen & 256 & 59.8 & 47.3\\
       \ding{204} & 336x & MLP & unfrozen &576 & 58.1 & 23.5\\
       \ding{205} & 448x & Conv2d & frozen & 256  & 60.8 & 28.4\\
      \bottomrule[1pt]
  \end{tabular}
}
  \label{tab:ablation_res_proj}
\vspace{-6mm}
\end{table*}

%% file: Tabs/tab1.tex
\begin{table}[t]
\centering
\caption{
\textbf{All training data.} \textbf{Cap.}:Captioning, \textbf{Ref.}:Referring (including REC, REG and Referring Tracking), \textbf{Det.}: Detection, \textbf{Track}:Tracking (including single object tracking (SOT) and multiple object tracking (MOT)), \textbf{Rea.}:Reasoning, \textit{Dia.}:Dialogue. * means that the data is only used in the SFT stage.
}
\scalebox{1.0}{
\begin{tabular}{l| p{20mm} l| p{60mm} l}
\toprule
\textbf{Task} & \textbf{Data} & \textbf{Size} & \textbf{Task description example} \\
\midrule
Cap. & LAION & 10M & -- \\
\midrule
\multirow{4}*{Ref.} & RefCOCO & 200K & Refer to the supplied image and box coordinates for a short response to the question. \\
\cmidrule{2-4}
& MeViS & 180K & Provide the trajectory coordinate of the target according to the sentence describes. \\
\midrule
\multirow{3}*{Det.} & Object365 & 3.2M & \multirow{2}*{\makecell{When submitting your answer, maintain the \\category:[xmin,ymin,xmax,ymax] structure \\
consistently.}} \\
& OpenImage & 800K & \\
& CrowdHuman & 20K & \\
\midrule
\multirow{7}*{Track} & GOT10K & 100K & \multirow{7}*{\makecell{For the trajectories included in the answer,\\ please use the format \texttt{<Idi>Frame t:}\\\texttt{[xmin,ymin,xmax,ymax]</Idi>.}}} \\
& LaSOT & 15K & \\
& MOT17 & 10K & \\
& Sompt & 5K & \\
& DanceTrack & 25K & \\
& SportsMOT & 20K & \\
& BDD100K & 120K & \\
\midrule
Rea. & VCR & 250K & In response, account for any relevant object locations, denoted by [x0,y0,x1,y1].\\
\midrule
\multirow{2}*{Dia.*} & LLaVA-Ins & 665K & -- \\
& Merlin-chat & 30K & \\
\bottomrule
\end{tabular}
}
\label{tab:all_data}
\end{table}

%% file: Tabs/ablation_task_prompt.tex
\begin{table}[t!]
  \centering
    \caption{\textbf{Comparisions between Merlin with (w) and without (w/o) the precise task description.} We minaly report the AO score of GOT10K and the average score of Future Reasoning.}
  \setlength{\tabcolsep}{2.0pt}
  \begin{tabular}{@{}p{.8em}@{}l cccc c c@{}}
    \toprule
     && \multicolumn{1}{c}{$Setting$} && \multicolumn{2}{c}{$Metrics$} \\
    \cmidrule{3-3}
    \cmidrule{5-6}
	 && Precise Task description && GOT10K & Future Rea \\
    \midrule
	  \rownumber{1}	   && w && 51.4 & 64.4 \\
	  \rownumber{2}	   && w/o && 28.4 & 62.4 \\
    \bottomrule
  \end{tabular}
\label{tab:task}
\end{table} 

%% file: Tabs/ablation_tcot.tex
\begin{table}[htbp]
\caption{\small{Ablation studies of trajectory CoT}}
\vspace{-1mm}
  \centering
  \resizebox{0.9\textwidth}{!}{
    \renewcommand\arraystretch{0.93}
    \renewcommand\tabcolsep{0.85pt}
    \begin{tabular}{c|ccc|c}
      \toprule[1pt]
       Exp & Pretrain data &SFT data    & Trajectory CoT   &Prediction Reasoning\\ 
    \hline
        \ding{202} &FPT-data& LLaVA-665k & w/o & 61.2 \\
       \ding{203} & FPT-data&FIT-data & w/o & 63.0 \\
       \ding{204} & FPT-data&FIT-data & w & \bf 64.4 \\
      \bottomrule[1pt]
  \end{tabular}
}
  \label{table:tcot}
\end{table}